\documentclass{article} 
\usepackage{iclr2015,times}
\usepackage{hyperref}
\usepackage{url}

\usepackage{amsfonts,amstext,amsmath,amssymb}
\usepackage{longtable}

\usepackage{graphicx}
\usepackage{caption}
\usepackage{subcaption}
\usepackage{wrapfig}

\usepackage{tikz}
\usetikzlibrary{bayesnet}

\usepackage{booktabs} \newcommand{\ra}[1]{\renewcommand{\arraystretch}{#1}}

\title{Simple image description generator via \\a linear phrase-based model}

\author{
R\'emi Lebret$^{*}$ \& Pedro O. Pinheiro\thanks{These two authors contributed equally to this work.}    \\
Idiap Research Institute, Martigny, Switzerland\\
\'Ecole Polytechnique F\'ed\'erale de Lausanne (EPFL), Lausanne, Switzerland\\
\texttt{remi@lebret.ch, pedro@opinheiro.com} \\
\And
Ronan Collobert\thanks{All research was conducted at the Idiap Research Institute, before Ronan Collobert joined Facebook AI Research.} \\
Facebook AI Research, Menlo Park, CA, USA\\
Idiap Research Institute, Martigny, Switzerland\\
\texttt{ronan@collobert.com}
}

\iclrfinalcopy 

\begin{document}

\maketitle

\begin{abstract}
Generating a novel textual description of an image is an interesting problem that connects computer vision and natural language processing.  
In this paper, we present a simple model that is able to generate descriptive sentences given a sample image. 
This model has a strong focus on the syntax of the descriptions.
We train a purely bilinear model that learns a metric between an image representation (generated from a previously trained Convolutional Neural Network) and phrases that are used to described them. The system is then able to infer phrases from a given image sample. Based on caption syntax statistics, we propose a simple language model that can produce relevant descriptions for a given test image using the phrases inferred. Our approach, which is considerably simpler than state-of-the-art models, achieves comparable results on the recently release Microsoft COCO dataset.

\end{abstract}

\section{Introduction}
\label{intro}
Being able to automatically generate a description from an image is a fundamental problem in artificial intelligent, connecting computer vision and natural language processing. The problem is particularly challenging because it requires to correctly recognize different objects in images and also how they interact.

Convolutional Neural Networks (CNN) have achieved state of the art results in different computer vision tasks in the last few years. More recently, different authors proposed automatic image sentence description approaches based on deep neural networks. All the solutions use the representation of images generated by CNN that was previously trained for object recognition tasks as start point. 

\citet{VinyalsTBE14} consider the problem in a similar way as a machine translation problem. The authors propose a encoder/decoder (CNN/LSTM networks) system that is trained to maximize the likelihood of the target description sentence given a training image. \citet{KirosSZ14} also consider a encoder/decoder pipeline, but uses a combination of CNN and LSTM networks for encoding and a language model for decoding.
\citet{Andrej2014} propose an approach that is a combination of CNN, bidirectional recurrent neural networks over sentences and a structured objective responsible for a multimodal embedding. They propose a second recurrent neural network architecture to generate new sentences.
Similar to the previous works, \citet{MaoXYWY14} and \citet{DonahueHGRVSD14} propose a system that uses a CNN to extract image features and a deep recurrent neural network for sentences. The two networks interact with each other in a multimodal common layer.

\citet{FangGISDDGHMPZZ14} propose a different approach to the problem that does not rely on recurrent neural networks. Their solution can be divided into three steps: (i) visual detector for words that commonly occur are trained using multiple instance learning, (ii) a set of sentences are generated using a Maximum-Entropy language-model and (iii) the sentences are re-ranked using sentence-level features and a proposed deep multimodal similarity model. 

This paper proposes a different approach to the problem. We propose a system that at the same time: (i) automatically generates a sentence describing a given scene and (ii) is relatively simpler than the recently proposed approaches. Our model shares some similarities with previously proposed deep approaches. For instance, we also use a pre-trained CNN to extract image features and we also consider a multimodal embedding. However, thanks to the phrase-based approach, we do not use any complex recurrent network for sentence generation.

We represent the ground-truth sentences as a collection of noun, verb and prepositional phrases. Each phrase is represented by the mean of the vector representation of the words that compose it. We then train a simple \emph{linear} embedding model that transform an image representation into a multimodal space that is common to the image and the phrases that are used to describe them. To automatically generate sentences in inference time, we (i) infer the phrases that correspond to the sample image and (ii) use a simple language model based on the statistics of the ground-truth sentences present in the corpus.

\section{Phrase-based model for image descriptions}
\label{model}

\begin{figure}[!t]
\centering
\includegraphics[height=5cm]{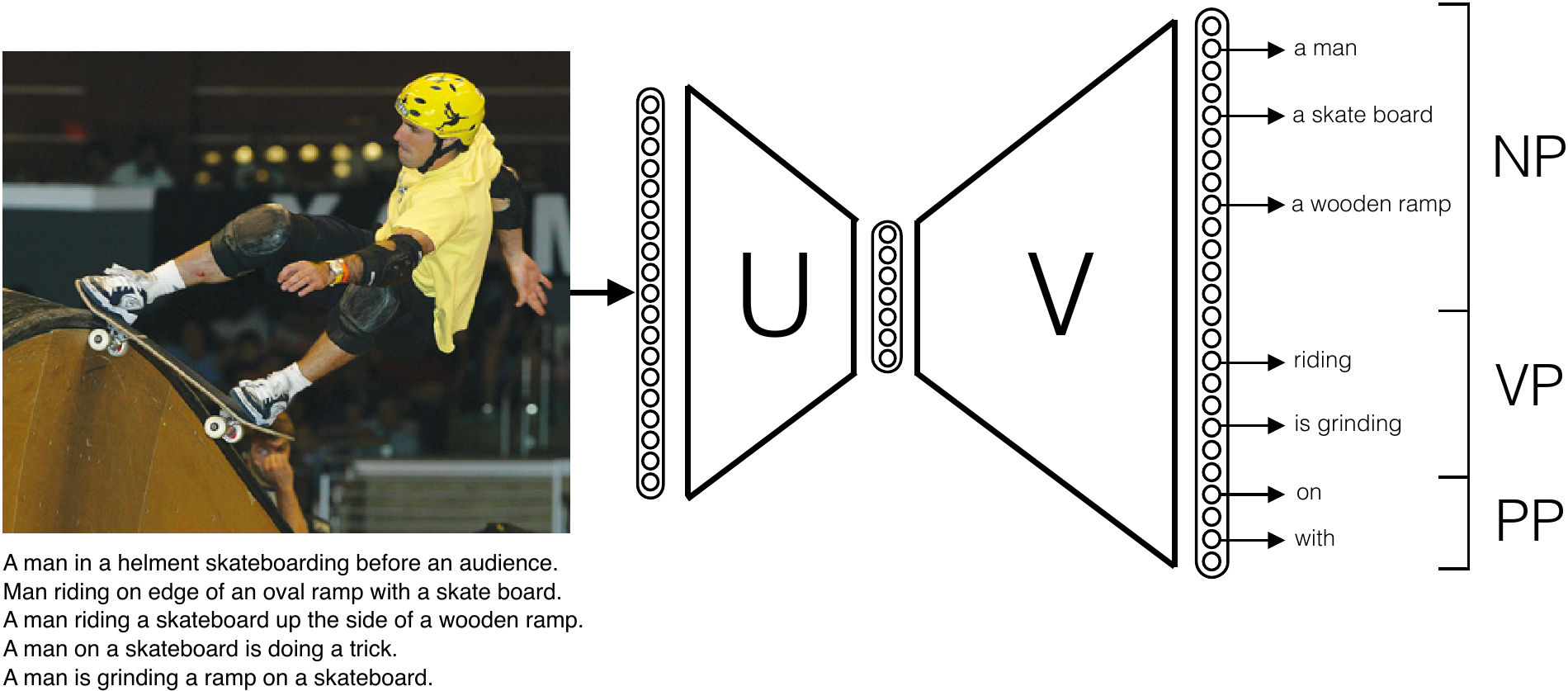}
\caption{Schematic illustration of our phrase-based model for image descriptions.}
\label{fig:schema}
\end{figure}

\subsection{Understanding structures of image descriptions}

The art of writing sentences can vary a lot according to the domain it is being applied. 
When reporting news or reviewing an item, not only the choice of the words might vary, but also the general structure of the sentence.
Sentence structures used for describing images can therefore be identified.

They possess a very distinct structure, usually describing the different objects present on the scene and how they interact between each other. This interaction among objects is described as actions or relative position between different objects.
The sentence can be short or long, but it generally respects this process.  
This statement is illustrated with the ground-truth sentence descriptions of the image in Figure \ref{fig:schema}.

\paragraph{Chunking-based approach}

All the key elements in a given image are usually described with a noun phrase (NP).
Interactions between these elements can then be explained using prepositional phrases (PP) or verb phrases (VP).
Describing an image is therefore just a matter of identifying these constituents to describe images.
We propose to train a model which can predict the phrases which are likely to be in a given image.

\paragraph{Phrase representations}

Noun phrases or verb phrases are often a combination of several words.
Good word vector representations can be obtained very quickly with many different recent approaches~\citep{Mikolov2013,Mnih2013,pennington2014glove,Lebret14b}.
\citet{MikolovICLR2013} also showed that simple vector addition can often produce meaningful results, such as \emph{king - man + woman $\approx$ queen}.
By leveraging the ability of these word vector representations to compose, representations for phrases are easily computed with an element-wise addition.

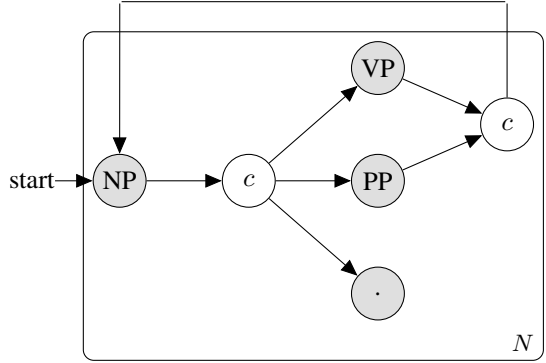
\begin{wrapfigure}{r}{0.5\textwidth}
  \vspace{-10pt}
  \begin{center}
%
%
%
%

\begin{tikzpicture}

  \node[obs]                  (np) {NP};
  \node[latent, right=1cm of np, yshift=0cm] (c1) {$c$};
  \node[obs, right=1cm of c1, yshift=1.5cm] (vp) {VP};
  \node[obs, right=1cm of c1, yshift=0cm] (pp) {PP};
  \node[obs, right=1cm of c1, yshift=-1.5cm] (period) {.};
  \node[latent, right=1 cm of pp, yshift=0.75cm] (c2)  {$c$} ; %

  \node[const, left=0.5 cm of np, yshift=-0.0cm] (start)  {start} ; %
  \node[const, above=1.25 cm of c2] (tmp2)  {} ; %
  \node[const, left= 5.1 cm of tmp2] (tmp3)  {} ; %

  \edge {start} {np} ; %
  \edge {np} {c1} ; %
  \edge {c1} {vp,pp,period} ; %
  \edge {vp,pp} {c2} ; %
  \myline {c2} {tmp2} ; %
  \myline {tmp2} {tmp3} ; %
  \edge {tmp3} {np} ; %

  \plate {} {(np)(vp)(pp)(period)(c1)(c2)} {$N$} ;

\end{tikzpicture}
  \end{center}
  \caption{The constrained language model for generating description given the predicted phrases for an image.}
  \vspace{-0pt}
  \label{fig:bayes}
\end{wrapfigure}

\paragraph{From phrases to sentence}

After identifying the most likely constituents of the image, we propose to use a statistical language model to combine them and generate a proper description. 
A general framework is defined to reduce the total number of combination and thus speed up the process for generating sentences. The constrained language model used is illustrated in Figure~\ref{fig:bayes}.
In general, a noun phrase is always followed by a verb phrase or a prepositional phrase, and both are then followed by another noun phrase. This process is repeated $N$ times until reaching the end of a sentence (characterized by a period). This heuristic is based on the analysis of syntax if the sentences (see Section~\ref{exp-results-setup}).

\subsection{A multimodal representation}

\paragraph{Image representations}
For the representation of images, we choose to use a Convolutional Neural Network. CNNs have been widely used for many different vision domains and are currently the state-of-the-art in many object recognition tasks. We consider a CNN that has been pre-trained for the task of object classification. We use a CNN solely to the purpose of feature extraction, that is, no learning is done in the CNN layers.

\paragraph{Learning of a common space for image and phrase representations}
Let $\mathcal{I}$ be the set of training images, $\mathcal{D}$ the set of all sentence descriptions for $\mathcal{I}$, $\mathcal{C}$ the set of all phrases occuring in $\mathcal{D}$, and $\theta$ the trainable parameters of the model.
$\mathcal{D}_i$ is the set of sentences describing a given image $i \in \mathcal{I}$, and $\mathcal{C}_d$ is the set of phrases which compose a sentence description $d \in \mathcal{D}_i$.
The training objective is to find the phrases $c$ that describe the images $i$ by maximizing the log probability:
\begin{equation}\label{eq:obj}
\sum_{i \in \mathcal{I}} \sum_{d \in \mathcal{D}_i} \sum_{c \in \mathcal{C}_d}  \log p(c | i)
\end{equation}

Each image $i \in \mathcal{I}$ is represented by a vector $\mathrm{x}_i \in \mathbb{R}^n$ thanks to a pre-trained CNN. 
Each phrase $c$ is composed of $K$ words $w$ which are represented by a vector $\mathrm{x}_{w} \in \mathbb{R}^m$ thanks to another pre-trained model for word representations.
A vector representation $\mathrm{z}_c$ for a phrase $c = \{w_1,\ldots,w_K\}$ is then calculated by averaging its word vector representations:
\begin{equation}
 \mathrm{z}_{c} = \frac{1}{K} \sum_{k=1}^K \mathrm{x}_{w_k}\,.
\end{equation}
Vector representations for all phrases $c \in \mathcal{C}$ can thus be obtained to build a matrix $V = \left[ \mathrm{z}_{c_1}, \ldots, \mathrm{z}_{c_{|\mathcal{C}|}} \right] \in \mathbb{R}^{m \times |\mathcal{C}|}$ .
In general, $m \ll n$. An encoding function is therefore defined to map image representations $\mathrm{x}_i \in \mathbb{R}^n$ in the same vector space than phrase representations $\mathrm{z}_c \in \mathbb{R}^m$:
\begin{equation}
g_{\theta}(i) =  \mathrm{x}_i U\,,
\end{equation}
where $U \in \mathbb{R}^{n \times m}$ is initialized randomly and trained to encode images in the same vectorial space than the phrases used for their descriptions. 
Because representations of images and phrases are in a common vector space, similarities between a given image $i$ and all phrases can be calculated:
\begin{equation}
f_{\theta}(i) =  g_{\theta}(i) V\,,
\end{equation}
where $V$ is fine-tuned to incorporate other features coming from the images. 
By denoting $[f_{\theta}(i)]_j$ the score for the $j^{th}$ phrase, this score can be interpreted as the conditional probability $p(c = c_j | i, \theta)$ by applying a softmax operation over all the phrases:
\begin{equation}
p(c = c_j | i, \theta) =  \frac{ e^{[f_{\theta}(i) ]_j} }{ \sum_{k=1}^{|\mathcal{C}|} e^{ [f_{\theta}(i)]_k}}\,.
\end{equation}
In practice, this formulation is often impractical due to the large set of possible phrases $\mathcal{C}$.

\paragraph{Training with negative sampling}
With $\theta=\{U,V\}$ and a negative sampling approach, we instead minimize the following logistic loss function with respect to $\theta$:
\begin{equation}
\theta \mapsto  \sum_{i \in \mathcal{I}} \sum_{d \in \mathcal{D}_i} \sum_{c_{j} \in \mathcal{C}_d}  \Big( \log \Big(1 +e^{[f_{\theta}(i) ]_j}\Big) + \sum_{k=1}^N \log \Big(1 +e^{-[f_{\theta}(i) ]_k}\Big) \Big) \,.
\end{equation}
Thus the task is to distinguish the target phrase from draws from the noise distribution, where there are $N$ negative samples for each data sample. 
The model is trained using stochastic gradient descent.

\section{Experiments}
\label{exp-results}
\subsection{Experimental Setup}
\label{exp-results-setup}
\begin{figure}[!t]
\centering
\begin{subfigure}[b]{0.5\textwidth}
        \includegraphics[width=\textwidth]{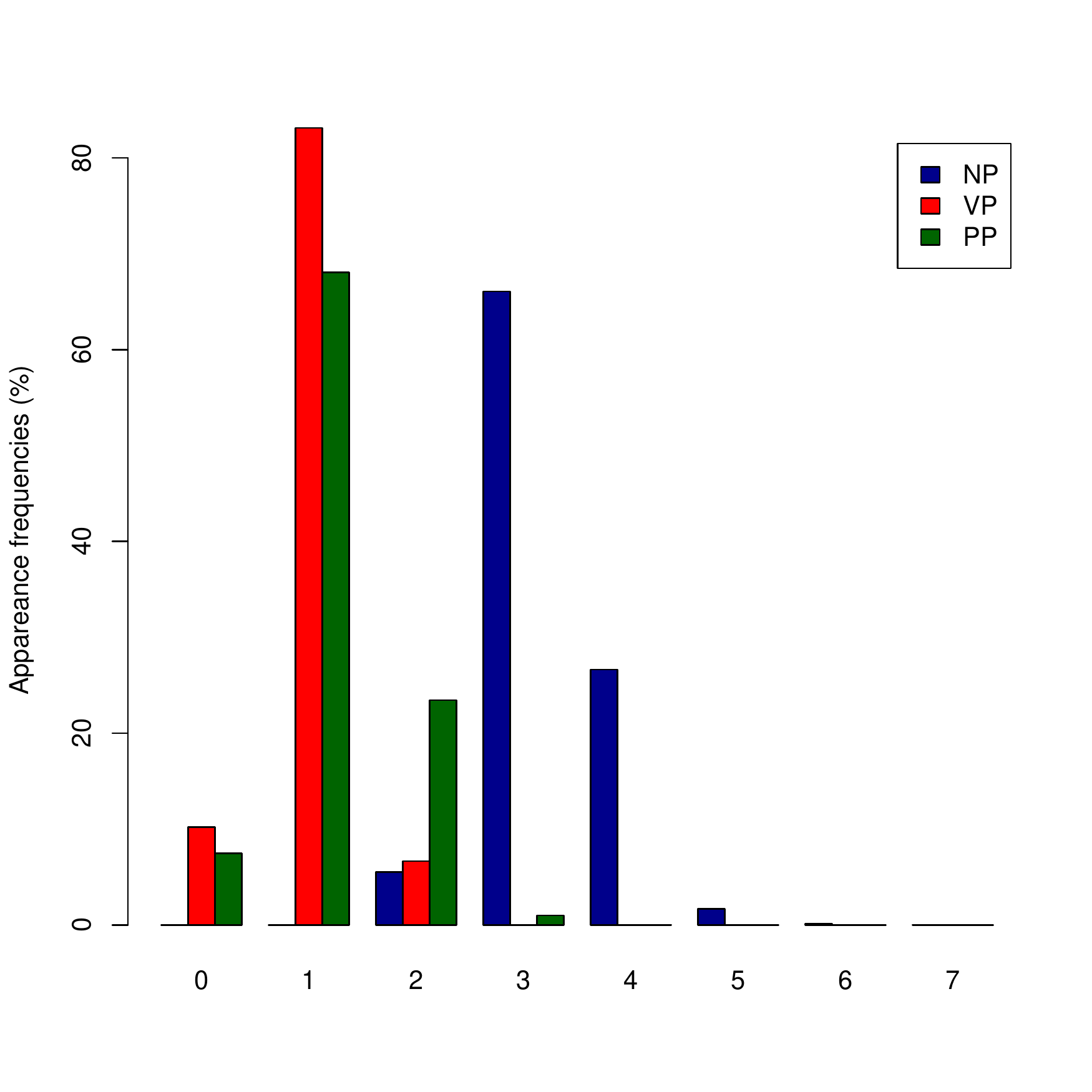}
        \caption{The number of phrases per sentence.}
\end{subfigure}%
~
\begin{subfigure}[b]{0.5\textwidth}
        \includegraphics[width=\textwidth]{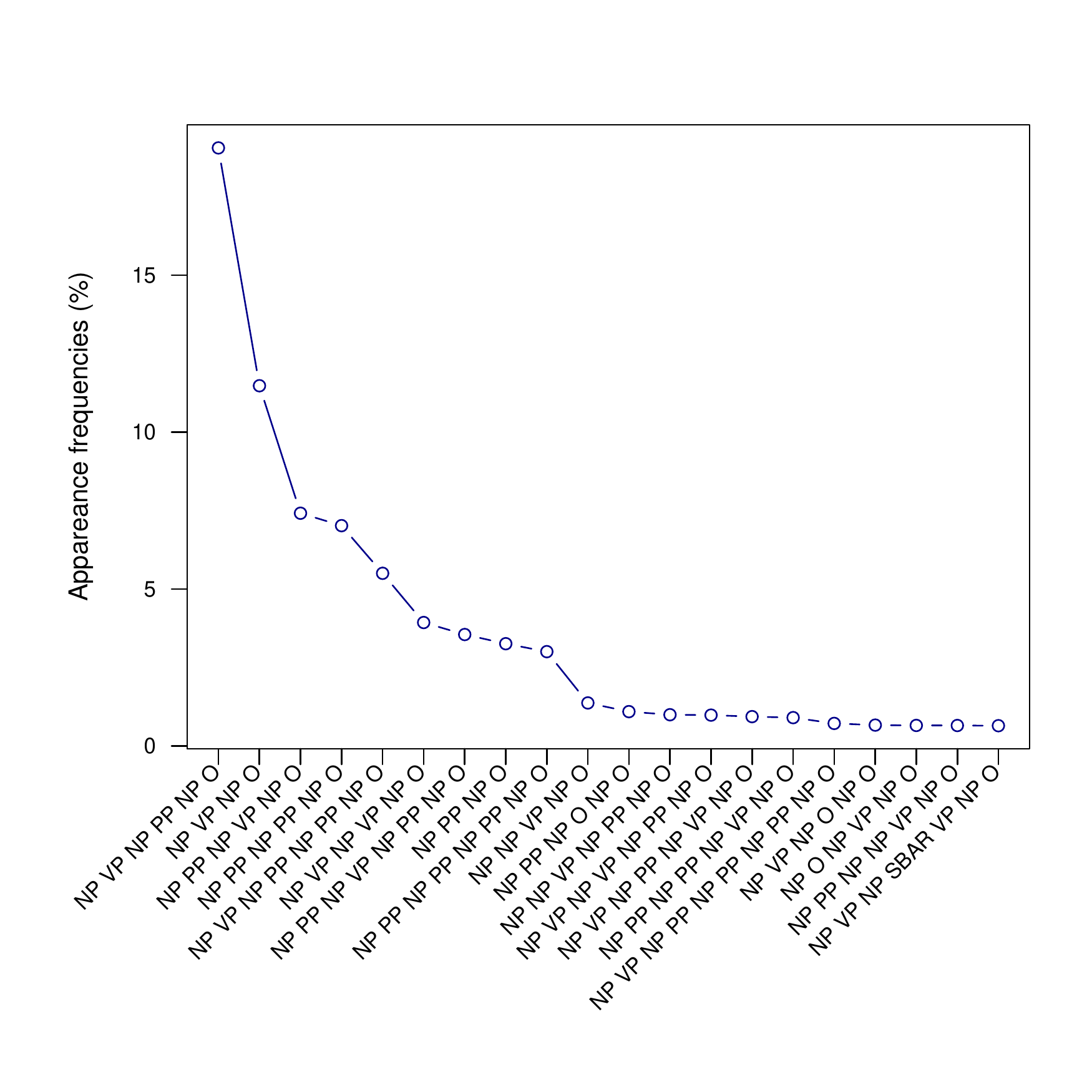}
        \caption{The 20 most frequent sentence syntactic structures.}
\end{subfigure}
\caption{Sentence structure statistics of COCO datasets.}
\label{fig:stats}
\end{figure}

\paragraph{Dataset} We validate our model on the recently proposed COCO dataset~\citep{mscoco2014}, which contains complex images with multiple objects. The dataset contains a total of 123,000 images, each of them with 5 human annotated sentences. The testing images has not yet been released.
We thus use two sets of 5,000 images from the validation images for validation and test, as in \citet{Andrej2014}\footnote{Available at \small\url{http://cs.stanford.edu/people/karpathy/deepimagesent/}}. We measure the quality of of the generated sentences using the popular, yet controversial, BLEU score~\citep{Papineni:2002}.

\paragraph{Feature selection} Following \citet{Andrej2014}, the image features are extracted using VGG CNN~\citep{Chatfield14}. This model generates image representations of dimension 4096 form RGB input images.
For sentence features, we extract phrases from the 576,737 training sentences with the SENNA software\footnote{Available at \small\url{http://ml.nec-labs.com/senna/}}. 
Statistics reported in Figure~\ref{fig:stats} confirm the hypothesis that image descriptions have a simple syntactic structure. A large majority of sentences contain from two to four noun phrases. Two noun phrases then interact using a verb or prepositional phrase.
Only phrases occuring at least ten times in the training set are considered. This results in 11,688 noun phrases, 3,969 verb phrases\footnote{Pre-verbal and post-verbal adverb phrases are merged with verb phrases.} and 219 prepositional phrases. 
Phrase representations are then computed by averaging vector representations of their words.
We obtained word vector representations from the Hellinger PCA of a word co-occurence matrix, following the method described in~\citet{Lebret14b}.
The word co-occurence matrix is built over the entire English Wikipedia\footnote{Available at \url{http://download.wikimedia.org}. We took the January 2014 version.}, with a symmetric context window of ten words coming from the 10,000 most frequent words. Words, and therefore also phrases, are represented in 400-dimensional vectors. 

\paragraph{Learning multimodal representation}
The parameters $\theta$ are $U \in \mathbb{R}^{4096 \times 400}$ and $V \in \mathbb{R}^{400 \times 15876}$. The latter is initialized with the phrase representations.
They are trained with $N=15$ negative samples and a learning rate set to 0.00025.

\paragraph{Generating sentences from the predicted phrases}
According to the statistics of ground-truth sentence structures, we set $N=\{2,3,4\}$. 
As nodes, we consider only the top twenty predicted noun phrases, the top ten predicted verb phrases and the top five predicted prepositional phrases.
A trigram language model is used for the transition probabilities between two nodes. 
The probability of each lexical phrase is calculated using the previous phrases, $p(c_j | c_{j-2}, c_{j-1})$, and the constraint described in Figure~\ref{fig:bayes}.
In order to reduce the number of sentences generated, we just consider the transitions which are likely to happen (we discard any sentence which would have a trigram transition probability inferior to 0.01). This thresholding also helps to discard sentences that are semantically incorrect.

\paragraph{Ranking generated sentences}
Our final step consists on ranking the sentences generated and choosing the one with the highest score as the final output. For each test image $i$, we generate a set of $M$ sentence candidates using the proposed language model. For each sentence $s_m$ ($m\in\{1,...,M\}$), we compute its vector representation $z_{s_m}$ by averaging the representation of the phrases $z_c \in V$ that make the sentence. The final score for each sentence $s_m$ is computed by doing a dot product between the sentence vector representation and the encoded representation of the sample image $i$:
\begin{equation}
f_{\theta}(i,m) =  g_{\theta}(i) z_{s_m}\,.
\end{equation}

The output of the system is the sentence which has the highest score. This ranking helps the system to chose the sentence which is closer to the sample image.

\subsection{Experimental Results} 
\begin{figure}[h]
  \begin{center}
    \includegraphics[width=\textwidth]{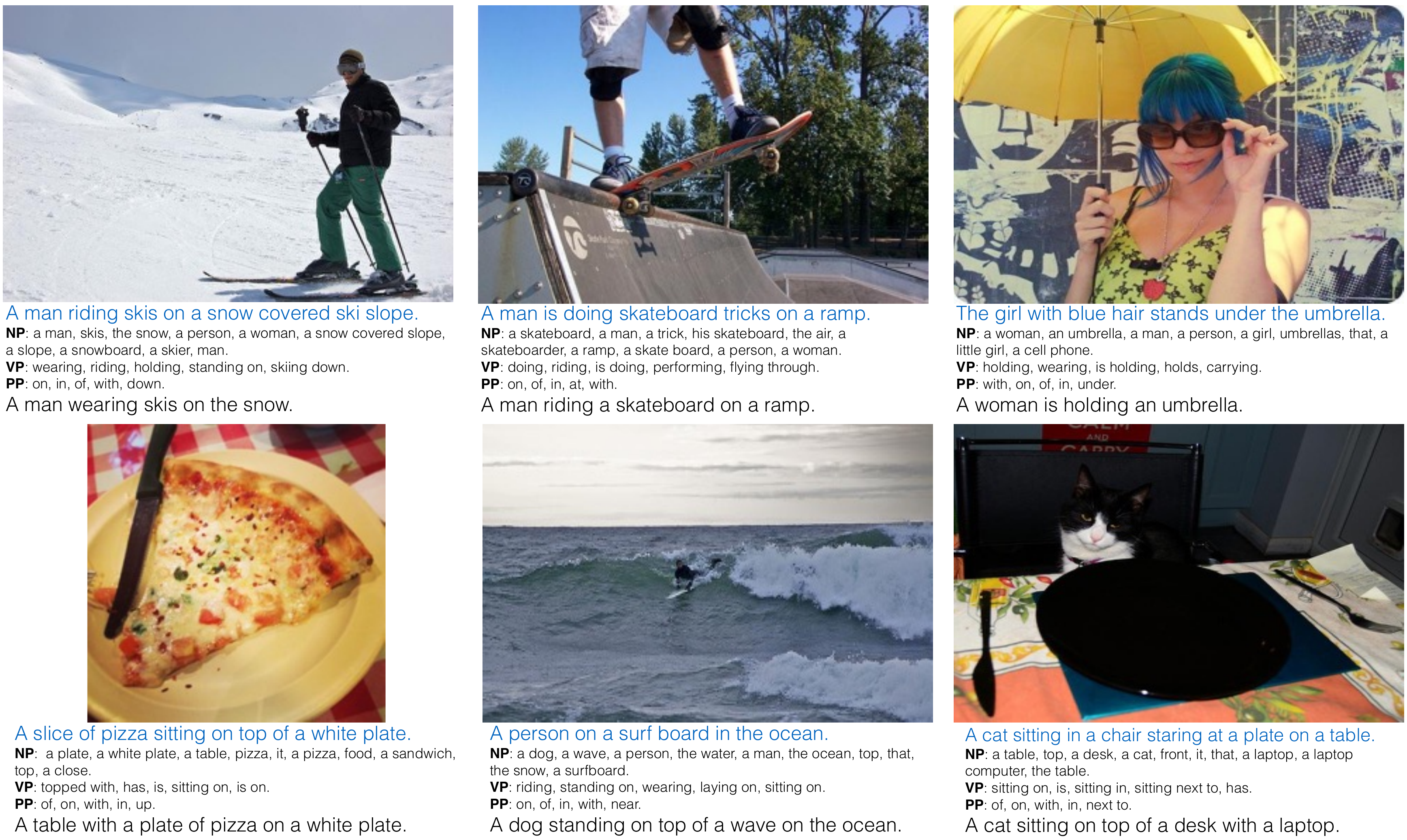}
  \end{center}
  \caption{Quantitative results for images on the COCO dataset.  Ground-truth annotation (in blue), the NP, VP and PP predicted from the model and generated annotation (in black) are shown for each image.  The two last are failure samples.}
  \label{fig:results}
\end{figure}

Table~\ref{tab:results} show our sentence generation results on the COCO dataset. 
BLEU scores are reported up to 4-grams. Human agreement scores are computed by comparing one of the ground-truth description against the others. 
For comparison, we include results from recently proposed models.
Although we use the same test set as in \citet{Andrej2014}, there are slight variations between the test sets chosen in other papers.
Our model gives competitive results at all N-gram levels. It is interesting to note that our results are very close to the human agreement scores. Examples of full automatic generated sentences can be found in Figure~\ref{fig:results}. 

\begin{table}
\ra{1.3}
\begin{center}
\begin{tabular}{@{}lccccc@{}}
\hline\toprule
{\bf Captioning Method} & \phantom{abc} & B-1 & B-2 & B-3 & B-4\\
\midrule
Human agreement & & $0.68$ & $0.45$ & $0.30$ & $0.20$\\
\midrule
\citet{Andrej2014} & & $0.57$ & $0.37$ & $0.19$ & -\\
\citet{VinyalsTBE14} & & $0.67$ & - & - \\
\citet{DonahueHGRVSD14} & &  $0.63$ & $0.44$ & $0.30$ & $0.21$\\
\citet{FangGISDDGHMPZZ14} & & - & - & - & $0.21$\\
Our model & & $0.70$ & $0.46$ & $0.30$ & $0.20$\\
\bottomrule
\hline
\end{tabular}
\end{center}
\caption{Comparison between human agreement scores, state of the art models and our model on the COCO dataset. Note that there are slight variations between the test sets chosen in each paper.}
\label{tab:results}
\end{table}

\section{Conclusion and future works}
\label{conclusion}
In this paper, we propose a simple model that is able to automatically generate sentences from an image sample. Our model is considerably simpler than the current state of the art, which uses complex recurrent neural networks.
We predict phrase components that are likely to describe a given image and use a simple statistical language model to generate sentences. Our model achieves promising first results. Future works include apply the model to different datasets (Flickr8k, Flickr30k and final COCO version for benchmarking), do image-sentence ranking experiments and improve the language model used.

\subsubsection*{Acknowledgements}

This work was supported by the HASLER foundation through the grant ``Information and Communication Technology for a Better World 2020'' (SmartWorld).

\bibliography{iclr2015}
\bibliographystyle{iclr2015}

\end{document}